\documentclass[11pt]{article}

\usepackage[letterpaper,margin=0.90in]{geometry}
\usepackage[T1]{fontenc}
\usepackage[utf8]{inputenc}
\usepackage{lmodern}
\usepackage{microtype}
\usepackage{amsmath,amssymb,bm}
\usepackage{booktabs}
\usepackage{longtable}
\usepackage{tabularx}
\usepackage{array}
\usepackage{enumitem}
\usepackage[dvipsnames,table]{xcolor}
\usepackage{graphicx}
\usepackage{caption}
\usepackage{fancyhdr}
\usepackage{titlesec}
\usepackage{natbib}
\usepackage{xurl}
\usepackage[colorlinks=true,allcolors=MidnightBlue]{hyperref}

\definecolor{ink}{HTML}{152238}
\definecolor{accent}{HTML}{355C7D}
\definecolor{soft}{HTML}{EEF3F7}
\definecolor{rulegray}{HTML}{AAB6C1}

\hypersetup{
  pdftitle={Blog: Survey of Optimizers},
  pdfauthor={Ruoran Xu},
  pdfsubject={A survey of neural-network optimizers and training optimization}
}

\renewcommand{\headrulewidth}{0.3pt}
\renewcommand{\headrule}{\hbox to\headwidth{\color{rulegray}\leaders\hrule height \headrulewidth\hfill}}

\titleformat{\section}{\Large\bfseries\color{ink}}{\thesection}{0.65em}{}
\titleformat{\subsection}{\large\bfseries\color{accent}}{\thesubsection}{0.6em}{}
\titlespacing*{\section}{0pt}{1.3em}{0.55em}
\titlespacing*{\subsection}{0pt}{1.0em}{0.35em}

\setlist[itemize]{leftmargin=1.35em,itemsep=0.15em,topsep=0.25em}
\setlist[enumerate]{leftmargin=1.55em,itemsep=0.15em,topsep=0.25em}
\newcolumntype{Y}{>{\raggedright\arraybackslash}X}
\newcolumntype{P}[1]{>{\raggedright\arraybackslash}p{#1}}

\newcommand{\work}[1]{\textit{#1}}
\newcommand{\AdamW}{\textsc{AdamW}}
\newcommand{\Muon}{\textsc{Muon}}
\newcommand{\SOAP}{\textsc{SOAP}}
\newcommand{\Shampoo}{\textsc{Shampoo}}
\newcommand{\SPlus}{\textsc{SPlus}}

\title{\vspace{-1.7em}\textbf{\color{ink}Blog: Survey of Optimizers}}
\author{Ruoran Xu}
\date{Literature current through 28 August 2026}

\begin{document}
\maketitle
\vspace{-1.4em}

\begin{abstract}
\noindent
Neural-network optimization in 2025--2026 is not well described as a parade of new Adam variants. The design unit has expanded: from coordinates to matrices and layers, from a fixed training horizon to a policy over time, and from a mathematical update rule to a state representation that must survive sharding and low precision. This blog survey organizes recent optimizers and training-optimization methods along four largely independent axes: \emph{temporal estimation}, \emph{update geometry}, \emph{horizon management}, and \emph{representation and systems}. It connects the spectral normalization of \Muon, the historical matrix statistics of \Shampoo{} and \SOAP, adaptive and hybrid matrix methods, memory-reduced optimizers, schedule-free training, small-batch corrections, and quantized optimizer states. The central empirical conclusion is deliberately non-triumphal: matrix-aware methods are a real advance, but there is no context-free replacement for \AdamW. Rankings change with scale, data-to-parameter ratio, batch size, schedule, parameter partition, tuning budget, and whether the objective is tokens, FLOPs, wall-clock time, or memory. The practical consequence is a compositional view of optimizer design and a stricter evaluation protocol.
\end{abstract}

\clearpage
\tableofcontents
\clearpage

\section{The optimizer question has changed}

For most practitioners, the optimizer question used to be a short menu: SGD with momentum, Adam, or \AdamW. The choice mattered, but the rest of the training stack could be discussed separately. That separation is breaking down. A modern optimizer may decide how long to remember a gradient, which matrix norm defines a small update, whether a low-rank subspace should carry state, how learning-rate decay is emulated when the stopping time is unknown, and how moment estimates are encoded in 8, 4, or even fewer bits.

This change is easiest to miss when every proposal is treated as one more named optimizer. The useful unit of analysis is instead an \emph{optimizer stack}. For a parameter block $\theta_t$ and stochastic gradient $g_t$, write schematically
\begin{align}
z_t &= \mathcal{T}_t(g_{1:t}), &
u_t &= \mathcal{P}_t(z_t;S_t), \nonumber\\
\widehat{u}_t &= \mathcal{Q}_t(u_t), &
\theta_{t+1} &= \mathcal{R}_t(\theta_t,\widehat{u}_t,\eta_t,\lambda_t).
\label{eq:stack}
\end{align}
Here $\mathcal{T}_t$ is a temporal estimator such as momentum or a variance-reduced recursion; $\mathcal{P}_t$ sets the geometry through diagonal, Kronecker, spectral, or low-rank structure; $\mathcal{Q}_t$ is the stored and communicated representation; and $\mathcal{R}_t$ includes learning-rate policy, averaging, and regularization. Two optimizers with different names can modify the same box. Conversely, a single implementation can quietly combine ideas from all four.

\begin{table}[t]
\centering
\caption{A functional map of the recent optimizer landscape. Persistent state is approximate and omits implementation-dependent master weights.}
\label{tab:map}
\small
\rowcolors{2}{soft}{white}
\begin{tabularx}{\textwidth}{P{0.15\textwidth}Y Y P{0.19\textwidth}}
\toprule
\textbf{Axis} & \textbf{Question} & \textbf{Representative mechanisms} & \textbf{Typical failure mode} \\
\midrule
Temporal estimation & Which past gradients should influence the next step? &
Fast/slow EMAs, recursive variance reduction, agreement masks, spike resets &
Stale or noisy history; sensitivity to $\beta$ and batch size \\
Geometry & What counts as an equally sized update? &
Diagonal RMS scaling, Kronecker whitening, polar factors, row/column normalization &
Costly state; unstable scaling; inappropriate treatment of non-matrix tensors \\
Horizon & How should progress be exposed over time? &
Cosine or WSD schedules, cooldowns, averaging, schedule-free interpolation &
Unknown stopping time; endpoint quality that hides pathwise progress \\
Representation & What can be stored and moved? &
Block statistics, low-rank moments, stateless transforms, FP8/INT8/2--4 bit states &
Quantization staleness; changing-basis errors; communication bottlenecks \\
\bottomrule
\end{tabularx}
\end{table}

This review focuses on work released from January 2025 through August 2026, with a small number of late-2024 precursors needed to make the lineage intelligible. It is a field guide rather than a claim of exhaustive bibliometrics. The emphasis is large-scale language-model pretraining because that is where the interaction between convergence, memory, and systems is most visible, but most mechanisms are more general.

\subsection{How to read the evidence}

Optimizer papers are unusually easy to compare unfairly. A result can be true within its protocol and still reverse under another horizon or batch size. Throughout this article, claims should be read at one of three levels:

\begin{itemize}
  \item \textbf{Scaled evidence}: experiments at roughly billion-parameter scale or long multi-billion-token horizons. This is valuable systems evidence, not universality.
  \item \textbf{Controlled evidence}: careful sweeps, ablations, or multiple scales designed to isolate components. These studies may use smaller models precisely because control is expensive.
  \item \textbf{Emerging evidence}: a recent preprint, limited scale, or a method whose independent replication is not yet mature.
\end{itemize}

Numbers below are attributed to the corresponding study and are not presented as directly comparable league-table scores. Faster may mean fewer tokens to a validation loss, fewer FLOPs, shorter wall-clock time, or a larger feasible batch. These are different claims.

\section{Foundations: what an optimizer chooses}

The conventional update $\theta_{t+1}=\theta_t-\eta_t g_t$ hides almost every choice that matters in contemporary training. An optimizer chooses an estimator of direction, a metric for measuring the size of that direction, a temporal policy, a regularizer, a representation for persistent state, and a distribution strategy. The recent literature becomes much easier to parse when these choices are separated.

\subsection{Direction, metric, and representation}

Let $G_t\in\mathbb{R}^{m\times n}$ denote the gradient of a matrix-valued parameter block. A general preconditioned step can be written
\begin{equation}
\Delta W_t=-\eta_t A_t\,\Phi_t(G_{1:t})\,B_t,
\label{eq:general-matrix-step}
\end{equation}
where $\Phi_t$ is a temporal estimator and $A_t,B_t$ encode left and right geometry. Several familiar optimizers are special cases.

\begin{itemize}
  \item In momentum SGD, $\Phi_t$ is an exponential moving average and $A_t=B_t=I$.
  \item In Adam, the operator is diagonal in the coordinate basis; each scalar receives its own running scale.
  \item In Shampoo, $A_t$ and $B_t$ are inverse roots of accumulated row and column covariance matrices.
  \item In Muon, Eq.~\eqref{eq:general-matrix-step} is better interpreted through a constrained linear problem: the momentum matrix is mapped approximately to its polar factor, which is the steepest direction under a spectral-norm trust region.
  \item In low-rank methods, $A_t$ and $B_t$ may be implicit projection operators, with adaptive statistics stored only in the projected coordinates.
\end{itemize}

These operations can produce similar-looking normalized matrices while answering different questions. A preconditioner estimates how the objective stretches space. A normalizer enforces a chosen update magnitude. A projection decides where information is retained. A quantizer decides how accurately the retained information survives. Confusing these roles is a recurring source of overbroad claims.

The distinction also clarifies the word \emph{adaptive}. At least four meanings occur in current papers:
\begin{enumerate}
  \item \textbf{coordinate adaptation}: a different effective step size for each scalar, as in Adam;
  \item \textbf{matrix adaptation}: a basis or whitening transform learned from matrix statistics;
  \item \textbf{temporal adaptation}: a changing mixture of recent and old gradients, or resets triggered by instability;
  \item \textbf{resource adaptation}: precision, rank, or state granularity selected according to layer and time.
\end{enumerate}
Calling all four adaptive without qualification obscures which failure mode is being addressed.

\subsection{From Euclidean steepest descent to norm-aware updates}

For a differentiable objective $f$ and a norm $\|\cdot\|$, the linearized steepest-descent direction solves
\begin{equation}
d^\star=\arg\min_{\|d\|\leq 1}\langle \nabla f(\theta),d\rangle.
\label{eq:lmo}
\end{equation}
Under the Euclidean norm, $d^\star$ is the normalized negative gradient. Under an $\ell_\infty$ constraint, the solution is a sign vector. For a matrix constrained in spectral norm, the solution is the negative polar factor $-UV^\top$ of the gradient. This gives a common language for sign methods, Lion \citep{chen2023lion}, Muon, and the norm-designed family developed by Scion \citep{pethick2025scion}.

The norm view is useful, but it does not make every norm-based method a second-order optimizer. Equation~\eqref{eq:lmo} changes the geometry of the trust region without estimating curvature. A second-order or natural-gradient method instead uses an operator $C_t$ intended to approximate a Hessian, generalized Gauss--Newton matrix, Fisher matrix, or gradient covariance:
\begin{equation}
d_t=-(C_t+\lambda I)^{-1}g_t.
\label{eq:second-order}
\end{equation}
K-FAC approximates layerwise Fisher blocks by Kronecker products \citep{martens2015kfac}; Shampoo builds Kronecker statistics from gradients \citep{gupta2018shampoo}; Sophia uses a lightweight diagonal curvature estimate \citep{liu2023sophia}. Their computational compromises differ, but all try to use information beyond the instantaneous direction.

A third category is \emph{whitening}. If $L_t$ and $R_t$ estimate row and column second moments, $L_t^{-1/4}G_tR_t^{-1/4}$ approximately balances the covariance of the transformed gradient. Whitening can resemble both preconditioning and spectral normalization, yet it additionally changes variance across time and coordinates. The 2025 component studies show that this variance-adaptation role is empirically essential \citep{frans2025matters}. It is therefore better to ask which part of a whitening transform matters than to label the entire family second order.

\subsection{The optimizer state as a statistical object}

Persistent state is not merely an implementation detail. It is a compressed summary of the gradient history. For an EMA
\begin{equation}
s_t=\beta s_{t-1}+(1-\beta)x_t,
\end{equation}
the nominal memory in steps is approximately $(1-\beta)^{-1}$. But the useful memory depends on the sampling unit. If a step contains $B$ tokens, the state remembers roughly $B/(1-\beta)$ tokens. Changing the batch while holding $\beta$ fixed changes the statistical estimator even if the update code is identical.

State has three distinct dimensions:
\begin{itemize}
  \item \textbf{support}: which subspace, block, row, column, or coordinate receives a statistic;
  \item \textbf{timescale}: how past observations are weighted;
  \item \textbf{numerical representation}: the precision, scale, and quantizer used to store it.
\end{itemize}
Adam-mini changes support, AdEMAMix changes timescale, and SOLO changes representation. LDAdam changes all three because the subspace itself moves. This decomposition helps explain why two methods with similar byte counts may behave very differently.

\subsection{Parameter classes are not interchangeable}

Transformers contain several parameter classes with different shapes and symmetries: token embeddings, attention projections, gated MLP matrices, normalization scales, output heads, and sometimes mixture-of-experts routers. A matrix transform defined for a dense hidden weight does not automatically make sense for a vector or an embedding table. Practical Muon stacks therefore route many non-hidden parameters to AdamW.

This routing decision can materially affect results. Embeddings often have sparse or highly non-uniform row access; output heads see high variance; normalization parameters are one-dimensional; router matrices have permutation structure tied to experts. A symmetry-compatible construction formalizes the principle that an update should commute with the transformations under which the represented function is unchanged \citep{lau2026symmetry}. The result is not one universal matrix rule but a layerwise stack: two-sided spectral updates for some blocks, one-sided or row-aware transforms for others, and coordinatewise methods where the symmetry is coordinate-specific.

\begin{table}[t]
\centering
\caption{Three questions that distinguish methods often grouped under geometric optimization.}
\label{tab:geometry-questions}
\small
\rowcolors{2}{soft}{white}
\begin{tabularx}{\textwidth}{P{0.20\textwidth}Y Y Y}
\toprule
\textbf{Question} & \textbf{Spectral / norm-based} & \textbf{Whitening / covariance} & \textbf{Curvature / natural gradient} \\
\midrule
What is estimated? & Often no curvature; a direction is normalized under a chosen norm & Historical row/column or rotated-basis second moments & Fisher, Hessian, Gauss--Newton, or a structured surrogate \\
What is controlled? & Singular values or an induced update norm & Covariance and variance of transformed gradients & Predicted objective or distributional change \\
Representative methods & Muon, Scion, SoftMuon & Shampoo, SOAP, SPlus, COSMOS & K-FAC, Sophia, structured Fisher methods \\
Main cost & Matrix transforms and shape policy & Matrix state plus inverse roots/eigenbases & Curvature estimation, damping, inversion, communication \\
\bottomrule
\end{tabularx}
\end{table}

\subsection{Five budgets, five different winners}

An optimizer can be evaluated under at least five budgets:
\begin{enumerate}
  \item \textbf{tokens}: how many training examples reach a target loss;
  \item \textbf{gradient evaluations}: especially relevant when probing, sharpness-aware, or multi-evaluation steps are used;
  \item \textbf{FLOPs}: including orthogonalization, eigendecomposition, and state refresh;
  \item \textbf{wall-clock time}: including kernels, communication, synchronization, and checkpoint recovery;
  \item \textbf{memory}: including the opportunity to increase batch, context length, or model size.
\end{enumerate}
Per-step speed and per-token convergence can move in opposite directions. A matrix method may save 20\% of tokens but add 5\% to step time; a low-memory method may match Adam per token yet win decisively because it permits a fourfold larger batch. Conversely, a theoretically efficient preconditioner can lose in wall-clock time when a matrix operation serializes a distributed step.

For this reason, the remainder of the survey treats the optimizer as a budgeted policy. The relevant question is not only whether an update is mathematically attractive, but whether its statistics, numerical representation, and implementation convert that attraction into the resource metric that constrains the run.

\section{Why AdamW remains the reference point}

\AdamW{} combines exponential first and second moments with decoupled weight decay \citep{kingma2014adam,loshchilov2017adamw}. Ignoring bias correction for compactness,
\begin{align}
m_t &= \beta_1m_{t-1}+(1-\beta_1)g_t,\\
v_t &= \beta_2v_{t-1}+(1-\beta_2)g_t^2,\\
\theta_{t+1} &= (1-\eta_t\lambda)\theta_t
  -\eta_t\frac{m_t}{\sqrt{v_t}+\epsilon}.
\end{align}
Its continuing strength is not only diagonal adaptivity. \AdamW{} is a mature engineering contract: tensorwise behavior is predictable, sharding is routine, low-precision implementations are available, and years of recipes encode useful priors. A new method must beat this \emph{total risk}, not only a loss curve.

Recent work attacks three weaknesses of the Adam recipe: its temporal statistics may forget useful old gradients, its current gradient can interact awkwardly with its current denominator, and a second moment per scalar is expensive.

\subsection{More careful use of gradient history}

\work{AdEMAMix} adds a slow exponential average to the usual fast momentum, motivated by the observation that gradients from tens of thousands of steps ago can still be useful \citep{pagliardini2024ademamix}. In its largest reported comparison, a 1.3B model trained on 101B tokens matched an \AdamW{} model trained on 197B tokens. The result is striking, but it is best read as evidence that a single momentum timescale is an avoidable restriction, not as a portable twofold guarantee.

\work{MARS} introduces scaled stochastic recursive momentum, using a variance-reduced gradient estimator that can be paired with AdamW-, Lion-, or Shampoo-like preconditioners \citep{yuan2024mars}. The conceptual contribution is modular: improve the temporal estimator without committing to one geometry. \work{Cautious Optimizers} takes a more local route, masking coordinates where the proposed momentum update disagrees in sign with the current gradient \citep{liang2024cautious}. \work{Cautious Weight Decay} applies the same agreement intuition to regularization, decaying only weights for which decay is locally compatible with the optimizer's update \citep{chen2025cwd}.

\work{SPAM} is motivated by rare gradient spikes that can be orders of magnitude larger than surrounding values \citep{huang2025spam}. It combines spike-aware clipping with periodic resets of the momentum state. Resetting sounds crude, but it identifies a real failure mode: a contaminated EMA can remain wrong long after the triggering event.

\work{ADOPT} isolates a theoretical dependency in Adam by preventing the current gradient from determining the very denominator used to normalize it; operationally, normalization and moment accumulation are reordered \citep{taniguchi2024adopt}. This yields optimal convergence guarantees across a broad range of second-moment decay values. Controlled pretraining comparisons nevertheless find that empirical performance can still depend on $\beta_2$ \citep{semenov2025benchmark}. The lesson is useful beyond ADOPT: removing a parameter from a theorem is not the same as removing it from tuning.

\work{Singularity-aware Adam} targets a different instability: gradient chattering near non-smooth regions rather than ordinary stochastic variance \citep{xu2026singularity}. Randomized directional probes test whether nearby loss changes remain coherent; when they do not, the method temporarily damps the Adam update. Unlike sharpness-aware training, which changes the update using an adversarially perturbed neighborhood \citep{foret2020sam}, the probes act as a diagnostic brake while retaining Adam's direction. This places the work beside clipping, agreement filters, and state resets as a robustness mechanism, rather than as a separate optimizer family. Its present evidence is emerging, with the clearest motivation in quantization-aware, hard-routing, and high-noise training.

\subsection{A closer look at temporal estimators}

The first moment in Adam is often described simply as momentum, but this description hides a modeling choice: the optimizer assumes that one exponential timescale is an adequate summary of useful history. AdEMAMix replaces this with a short and a long timescale. In simplified notation,
\begin{align}
m_t^{\mathrm{fast}} &= \beta_f m_{t-1}^{\mathrm{fast}}+(1-\beta_f)g_t,\\
m_t^{\mathrm{slow}} &= \beta_s m_{t-1}^{\mathrm{slow}}+(1-\beta_s)g_t,\\
\widetilde m_t &= m_t^{\mathrm{fast}}+\alpha_t m_t^{\mathrm{slow}},
\end{align}
with $\beta_s$ close to one and a schedule for introducing the slow component \citep{pagliardini2024ademamix}. The mixture can retain directions that are individually too weak to dominate a short EMA. This is plausible in pretraining, where a gradient direction associated with a rare feature may recur after many unrelated batches.

The benefit is not free. A very slow average can carry obsolete directions after the model enters a new regime, and its timescale interacts with batch size and training duration. A comparison that copies $\beta_s$ across token budgets may inadvertently compare different estimators. Reporting EMA half-lives in tokens would make such studies substantially clearer.

MARS attacks variance more directly. Its recursive estimator combines current and previous stochastic gradients before feeding a chosen preconditioner \citep{yuan2024mars}. A generic recursive form is
\begin{equation}
c_t=g_t+\gamma_t(g_t-g_{t-1}^{\mathrm{matched}}),
\end{equation}
where the second gradient is evaluated or approximated so that the correction reduces stochastic drift. The precise MARS instances differ, but the important abstraction is that temporal variance reduction and spatial preconditioning can be designed independently. MARS-AdamW, MARS-Lion, and MARS-Shampoo are therefore a small family rather than one fixed rule.

\subsection{Agreement filters, resets, and robustness}

Cautious optimization asks whether a momentum proposal remains locally supported by the current gradient. For a proposed update $u_t$ and gradient $g_t$, an elementwise mask
\begin{equation}
c_t=\mathbb{I}\{u_t\odot g_t>0\}
\end{equation}
removes coordinates whose signs disagree, with normalization used to preserve overall scale \citep{liang2024cautious}. The intervention is intentionally minimal. It can be attached to AdamW or Lion without constructing new state. Its value is best understood as a trust filter: momentum is allowed to accelerate a direction only while current evidence has not reversed.

This filter is not guaranteed to help every noisy coordinate. Under symmetric noise, sign disagreement may be evidence of variance rather than a true change of direction. Excessive masking can reduce the effective dimension of the step and alter weight-decay balance. The correct comparison must therefore log the active-mask fraction and the RMS of the post-mask update.

Cautious weight decay applies related logic to the regularizer \citep{chen2025cwd}. Decoupled decay is normally applied uniformly even when it opposes the data-gradient update. A sign-conditioned version treats regularization as another proposal whose local compatibility can be checked. This does not make weight decay an unbiased estimator of any fixed penalty; it turns decay into a state-dependent training policy. That distinction matters when interpreting generalization.

SPAM focuses on rare, extreme events. Its study reports gradient spikes up to roughly three orders of magnitude above typical values in some pretraining traces \citep{huang2025spam}. Spike-aware clipping limits the immediate damage, while periodic state reset limits how long the event contaminates the first and second moments. These two operations target different terms:
\begin{itemize}
  \item clipping bounds the observation entering the state;
  \item reset truncates the influence of observations already stored.
\end{itemize}
Sparse momentum then reduces memory by retaining state for only a subset of coordinates. The combined method spans robustness and memory, which is why a single headline such as stable Adam understates its design.

Stable-SPAM extends this line to aggressive low-bit training \citep{huang2025stablespam}. It tracks historical gradient norms and spike thresholds, normalizes the full matrix, and retains periodic resets. The reported result that a 4-bit run can outperform a higher-precision Adam baseline is protocol-specific, but it makes an important mechanistic point: numerical precision and gradient stability are coupled. Low precision can amplify a poorly scaled update; a stabilization rule can be more valuable than simply adding bits.

\subsection{ADOPT and the timing of normalization}

Adam's current squared gradient contributes immediately to $v_t$ and is then divided by $\sqrt{v_t}$. The numerator and denominator are therefore statistically coupled. ADOPT uses the previous second-moment estimate to normalize the current gradient and updates the state afterward \citep{taniguchi2024adopt}. A simplified ordering is
\begin{align}
\bar g_t &= \frac{g_t}{\sqrt{v_{t-1}}+\epsilon},\\
m_t &= \beta_1m_{t-1}+(1-\beta_1)\bar g_t,\\
v_t &= \beta_2v_{t-1}+(1-\beta_2)g_t^2.
\end{align}
This lag removes a dependency that complicates convergence analysis and yields an optimal nonconvex rate under weaker conditions. Yet the benchmark literature still finds practical sensitivity to $\beta_2$ \citep{semenov2025benchmark}. The parameter no longer decides whether the proof works, but it still decides which temporal variance estimate is useful.

This difference between a \emph{validity parameter} and a \emph{performance parameter} recurs throughout optimization. Damping may be required for numerical stability but still need tuning for speed. An inverse-root refresh interval may preserve convergence while changing wall-clock performance. A schedule-free method may remove the need to know the final horizon while retaining a learning-rate-like scale.

\section{The matrix-aware renaissance}

The most visible shift of 2025--2026 is from coordinatewise scaling to matrix-aware update geometry. Neural-network parameters are not naturally a bag of unrelated scalars: dense weights, attention projections, embeddings, experts, and convolutional kernels have row/column structure. Matrix optimizers try to exploit that structure, but they do so in materially different ways.

\subsection{Whitening: Shampoo, SOAP, and stable descendants}

\Shampoo{} accumulates left and right gradient statistics for a matrix $G_t\in\mathbb{R}^{m\times n}$ and applies matrix inverse roots \citep{gupta2018shampoo}. In schematic form,
\begin{align}
L_t &= \beta L_{t-1}+(1-\beta)G_tG_t^\top,\\
R_t &= \beta R_{t-1}+(1-\beta)G_t^\top G_t,\\
\widetilde G_t &\approx L_t^{-1/4}G_tR_t^{-1/4}.
\end{align}
This can be read as Kronecker-factored whitening of historical gradients. Its costs are obvious: matrix states, eigendecompositions or inverse roots, and shape-dependent engineering.

\SOAP{} rotates the gradient into the evolving eigenbases of Shampoo-like statistics and runs an Adam-style update there \citep{vyas2024soap}. This decomposition matters conceptually. SOAP is not merely more second order than Adam; it combines a historical matrix coordinate system with coordinatewise adaptivity inside that system.

\SPlus{} revisits the same family with stability as the design target \citep{frans2025splus}. It uses a historical eigenbasis, an instantaneous normalization/sign-like operation, shape-aware scaling, and iterate averaging. Its authors report reaching Adam-level validation performance in an average of 44\% of the steps and 62\% of wall-clock time across their benchmark. \work{What Really Matters in Matrix-Whitening Optimizers?} provides the more important component-level interpretation: variance adaptation is critical, pure spectral normalization is effective but incomplete, and input-side preconditioning can recover much of SOAP's behavior with about 41\% of full SOAP's matrix-state memory in the studied setting \citep{frans2025matters}.

\subsection{Spectral steepest descent: what Muon is and is not}

\Muon{} applies momentum to each eligible matrix and approximately orthogonalizes the resulting update with a short Newton--Schulz iteration \citep{jordan2024muon,liu2025muonscalable}. If
\begin{equation}
B_t=\mu B_{t-1}+G_t,\qquad B_t=U\Sigma V^\top,
\end{equation}
the idealized transform is the polar factor
\begin{equation}
\operatorname{Polar}(B_t)=UV^\top.
\end{equation}
This replaces singular values by one while preserving singular directions. In the language of norm-constrained steepest descent, it is the linear minimization oracle associated with a spectral-norm constraint \citep{bernstein2024norm,pethick2025scion}.

That description also prevents a common category error. Muon is \textbf{matrix-aware}, but it is not a classical second-order method: it does not estimate a Hessian, and its Newton--Schulz iteration computes an approximate polar factor rather than a Newton step. Shampoo and SOAP use historical gradient covariance structure that can act as a preconditioner; Muon normalizes the current momentum matrix spectrally. Putting all three under second order erases the mechanism that should determine when they work.

The scalable Muon study reports matching an AdamW baseline with roughly 52\% of the training FLOPs under its fitted scaling-law protocol and demonstrates a 16B-parameter mixture-of-experts run \citep{liu2025muonscalable}. This is strong evidence that polar-style updates can survive serious scale. It is not evidence that every Muon implementation halves every training bill. Performance depends on how matrix and non-matrix tensors are separated, how shape scaling is applied, and how update RMS is aligned with the baseline.

\work{Scion} gives the norm-constrained view a broader vocabulary: choose a norm appropriate to a parameter block, compute its linear minimization oracle, and design scaling rules that transfer across width \citep{pethick2025scion}. This suggests a future in which embeddings, attention matrices, experts, and normalization parameters need not share one geometry.

\subsection{Adaptivity after orthogonalization}

Pure polar normalization deliberately discards singular-value magnitude. Several methods try to restore just enough adaptivity without returning to full SOAP:

\begin{itemize}
  \item \work{COSMOS} applies SOAP-like treatment in a leading eigensubspace and Muon-like treatment to the residual \citep{liu2025cosmos}.
  \item \work{AdaMuon} tracks elementwise second moments of orthogonalized updates and stabilizes their signs, combining spectral normalization with Adam-like variance information \citep{si2025adamuon}.
  \item \work{NorMuon} normalizes rows after orthogonalization using neuron-level second moments. Its authors report 21.74\% greater training efficiency than Adam and 11.31\% over Muon in a 1.1B setup, while providing a sharded implementation \citep{li2025normuon}.
  \item \work{Mousse} first whitens with Shampoo-derived Kronecker statistics and then takes a polar step in the whitened coordinates. On 160M--800M language models, its study reports roughly 12\% fewer steps than Muon with about 3\% wall-clock overhead \citep{sun2026mousse}.
\end{itemize}

Other recent preprints alter the geometry still further. \work{MuonEq} equilibrates row and column scales before orthogonalization \citep{chang2026muoneq}; \work{Muown} imposes explicit row-norm control to reduce spectral-norm drift \citep{lion2026muown}; \work{Hyperball} constrains Frobenius norms of both weights and updates \citep{wen2026hyperball}; and a softsign relaxation interpolates between sign-like and magnitude-sensitive behavior to address terminal oscillation \citep{feoktistov2026softsign}. These are emerging results, not a settled sequence of upgrades. Their collective message is more durable than any current ranking: the singular directions, row/column scales, and overall update norm are independent design choices.

\subsection{What controlled comparisons actually say}

Two broad comparisons make the current uncertainty unusually legible. One study evaluated ten optimizers from 0.1B to 1.2B parameters and from $1\times$ to $8\times$ a standard data-to-parameter training ratio \citep{wen2025fantastic}. Matrix-aware methods improved token efficiency by as much as roughly $1.4\times$ at small scale, but the gap narrowed to around $1.1\times$ at 1.2B. Muon was strongest at shorter ratios, while SOAP- and Kronecker-style methods could overtake it at longer ratios. Rankings at intermediate checkpoints did not reliably predict final rankings.

A separate benchmark of eleven methods, up to 720M parameters and 50B tokens, found AdEMAMix and MARS particularly strong in its larger-model, larger-batch setup, while Muon, SOAP, and distributed Muon variants could lead shorter runs \citep{semenov2025benchmark}. It also found schedule interactions: cosine was broadly strong, but Muon sometimes preferred a warmup-stable-decay policy.

These results do not cancel one another. They expose a basic fact:
\begin{quote}
\textbf{An optimizer winner is a statement about a protocol, not an intrinsic property of a name.}
\end{quote}

At even larger scale, a study of multi-billion-parameter models and trillion-token workloads stabilizes SOAP for large batches, matches update RMS across methods, and implements exact layerwise distributed optimizer operations \citep{khona2026soapmuon}. SOAP and Muon outperform AdamW in the tested regimes, with a SOAP variant slightly ahead. The systems work is part of the scientific claim: an algorithm whose communication cannot be hidden or whose matrix operations are approximated differently is a different practical optimizer.

\section{Dissecting matrix optimizers}

The names in the matrix-optimizer family are multiplying faster than the underlying mechanisms. A more stable taxonomy separates five operations: momentum formation, basis choice, singular-value transformation, variance adaptation, and output scaling.

\begin{table}[t]
\centering
\caption{Mechanism-level comparison of representative matrix-aware methods. A historical basis depends on accumulated statistics rather than only the current momentum.}
\label{tab:matrix-mechanisms}
\scriptsize
\rowcolors{2}{soft}{white}
\begin{tabularx}{\textwidth}{P{0.11\textwidth}P{0.15\textwidth}P{0.16\textwidth}P{0.17\textwidth}P{0.16\textwidth}Y}
\toprule
\textbf{Method} & \textbf{Basis} & \textbf{Spectrum treatment} & \textbf{Variance adaptation} & \textbf{Persistent matrix state} & \textbf{Primary design goal} \\
\midrule
Muon & Current momentum singular vectors & Approximately sets singular values to one & None in basic form & Momentum matrix & Cheap spectral steepest descent \\
Shampoo & Historical row/column statistics & Inverse-root whitening & Implicit in accumulated covariance & Left and right factors & Structured preconditioning \\
SOAP & Shampoo eigenbasis & Adam update in rotated coordinates & Elementwise in rotated basis & Factors plus Adam-style moments & Strong per-step adaptation \\
SPlus & Historical eigenbasis & Instantaneous bounded normalization & Sign-like / averaging based & Historical basis and auxiliary state & Stable, cheaper whitening \\
COSMOS & Leading historical subspace plus residual & SOAP in leading subspace, Muon outside & Concentrated in important subspace & Low-rank basis and moments & Memory-quality trade-off \\
AdaMuon & Orthogonalized-update coordinates & Polar transform plus sign stabilization & Elementwise second moment & Momentum and adaptive state & Restore coordinate variance \\
NorMuon & Polar update rows & Polar transform then row scaling & Neuron-wise second moment & Momentum plus row statistics & Balance neuron utilization \\
Mousse & Shampoo-whitened coordinates & Polar transform after whitening & Kronecker geometry & Shampoo-like factors & Anisotropic spectral step \\
\bottomrule
\end{tabularx}
\end{table}

\subsection{Muon's Newton--Schulz map}

Computing $UV^\top$ with an exact singular value decomposition every step would be expensive. Muon uses a small fixed number of Newton--Schulz iterations to approximate the polar factor. After normalizing a matrix $X_0$, a generic iteration has the form
\begin{equation}
X_{k+1}=aX_k+bX_kX_k^\top X_k+cX_kX_k^\top X_kX_k^\top X_k,
\label{eq:ns5}
\end{equation}
with coefficients chosen so singular values move rapidly toward one over a useful interval. Implementations differ in coefficients, normalization, number of iterations, and arithmetic precision. Those details affect both the approximation and kernel performance.

Several practical points follow.
\begin{itemize}
  \item \textbf{Rectangularity matters.} For a highly rectangular matrix, the polar factor has orthonormal columns or rows, not both. Implementations often transpose the matrix so the smaller dimension is treated favorably.
  \item \textbf{Finite iterations preserve spectral information.} A five-step approximation is not an exact sign of the singular values. The input condition number and stable rank influence the output.
  \item \textbf{Scale must be reintroduced.} A polar factor has a shape-dependent Frobenius norm. Without a shape rule, the same global learning rate produces different RMS updates for matrices of different aspect ratios.
  \item \textbf{Non-matrix parameters need a fallback.} Most stacks retain AdamW for embeddings, vectors, or selected heads, making the reported method a hybrid from the start.
\end{itemize}

MuonEq exploits the second point. It normalizes rows or columns before finite-step orthogonalization, improving the spectrum seen by Newton--Schulz without building a full whitening preconditioner \citep{chang2026muoneq}. The interpretation is useful: equilibration is a zeroth-order surrogate for whitening, and finite-step polar computation is itself a tunable part of the optimizer.

\subsection{Shape scaling, weight norms, and angular learning rate}

For a scale-invariant parameter block, the functional change induced by an update depends on the angle between the old and new weight matrices. If $W$ and $\Delta W$ are approximately orthogonal in Frobenius inner product, the angular step is roughly
\begin{equation}
\Delta\phi \approx \frac{\|\Delta W\|_F}{\|W\|_F}.
\end{equation}
Weight decay influences $\|W\|_F$ and therefore implicitly controls angular learning rate. This explains why optimizer comparisons can change dramatically with weight-decay conventions even when the data-gradient update is unchanged.

Hyperball makes the norm control explicit. It fixes or targets the Frobenius norms of weight matrices and optimizer updates, decoupling angular progress from the equilibrium created by constant decoupled decay \citep{wen2026hyperball}. Its reported 20--30\% token-equivalent advantage over weight-decay baselines on models up to 1.2B is evidence that norm dynamics are not a peripheral regularization detail.

Muown decomposes each matrix into row magnitudes and directions \citep{lion2026muown}. The row-magnitude vector receives an $\ell_\infty$-aligned update while Muon acts on the directional component. This addresses a diagnosed drift in spectral norm driven by row magnitudes. AngularMuown subsequently shows that the unnormalized directional parameterization implicitly decays angular step size and makes that multiplier explicit \citep{hubler2026angularmuown}. These methods recast weight decay, normalization, and scheduling as different controls over radial and angular motion.

\subsection{Whitening is not one operation}

Suppose $G$ has singular value decomposition $U\Sigma V^\top$. Exact two-sided whitening maps $G$ toward $UV^\top$, which looks like spectral normalization. But historical whitening uses covariance estimated over time:
\begin{equation}
\mathbb{E}[GG^\top]\quad\text{and}\quad\mathbb{E}[G^\top G].
\end{equation}
The eigenvectors and eigenvalues of these expectations need not match those of the current gradient. The resulting transform does two things:
\begin{enumerate}
  \item it reduces anisotropy across matrix directions;
  \item it adapts to persistent variance across time.
\end{enumerate}

The component study of matrix-whitening methods isolates this distinction by comparing sign-like and variance-adapted pairs: Signum versus Adam, SPlus versus SOAP, and Muon versus AdaMuon \citep{frans2025matters}. In its controlled setup, spectral accuracy alone does not explain performance; Muon can produce a tighter singular-value spread while SOAP achieves stronger per-step progress. Variance adaptation explains much of the gap.

This finding changes how low-memory approximations should be designed. If the benefit came only from exact spectral normalization, approximating the polar factor would be the central task. If historical variance matters, then a cheap estimator of variance in a useful basis can be more valuable than a more accurate instantaneous polar transform. Input-side preconditioning and low-rank variance estimators become principled compromises rather than incomplete SOAP.

\subsection{Why hybrids proliferate}

The leading-subspace strategy in COSMOS assumes that the most important covariance structure is concentrated in a small eigenspace \citep{liu2025cosmos}. SOAP is used where that structure justifies adaptive state; Muon handles the residual without paying full matrix-memory cost. Mousse instead applies a full conceptual composition: first transform into a Kronecker-whitened coordinate system, then solve a spectral steepest-descent problem there \citep{sun2026mousse}. The trust region is anisotropic in the original coordinates.

AdaMuon and NorMuon add variance after the polar transform at different granularities \citep{si2025adamuon,li2025normuon}. Elementwise adaptation has the greatest expressive power and highest state cost. Rowwise adaptation assumes that neurons, rather than scalar coordinates, are the meaningful units of imbalance. This granularity is a testable modeling assumption.

SoftMuon changes a different component. Hard polar or sign-like updates discard magnitude and can oscillate near convergence. A temperature-controlled soft-sign map interpolates toward magnitude-sensitive steps, with an adaptive quantile rule controlling temperature \citep{feoktistov2026softsign}. The proposal treats terminal convergence as a geometry-relaxation problem rather than a reason to abandon norm-aware optimization.

\subsection{Communication can be an optimizer operation}

Dion reduces distributed communication by maintaining local momentum and synchronizing compact orthonormalized update information rather than full gradients \citep{ahn2025dion}. The goal is not simply to reduce optimizer-state bytes; it changes what is exchanged between devices while retaining synchronous semantics. Such a method should be evaluated on communication volume, overlap, and the discrepancy between device-local states, not only validation loss.

Large-scale SOAP and Muon implementations take another route: distribute optimizer work layer by layer, balance matrix operations across ranks, and hide communication behind model computation \citep{khona2026soapmuon}. Exactness matters because approximating an eigenspace, delaying a factor update, or reducing orthogonalization steps can change the optimization rule. Systems comparisons should therefore state:
\begin{itemize}
  \item which tensors are sharded and when they are gathered;
  \item whether inverse roots, QR factorizations, or Newton--Schulz steps are exact relative to the single-device algorithm;
  \item the refresh interval for matrix statistics;
  \item the precision of factorization and state;
  \item whether optimizer communication is exposed or overlapped.
\end{itemize}

\subsection{A decision tree inside the matrix family}

The matrix family can be navigated through four questions.

\begin{enumerate}
  \item \textbf{Is persistent covariance information affordable?} If no, start with Muon or a stateless transform. If yes, a Shampoo/SOAP-style method becomes plausible.
  \item \textbf{Is the horizon short or long?} Pure spectral normalization is often strong early; long horizons more often reward variance adaptation.
  \item \textbf{Is memory or step time the binding resource?} SOAP can be step-efficient but state-heavy; Muon is state-light but still needs matrix kernels; COSMOS and SPlus occupy intermediate points.
  \item \textbf{Can parameter classes be routed explicitly?} If the implementation forces one transform onto embeddings, heads, and hidden matrices, theoretical elegance can become practical mismatch.
\end{enumerate}

The result is not a total order. Muon is the cleanest low-cost test of whether matrix geometry helps. SOAP is a strong test of whether historical covariance and rotated-basis adaptivity justify cost. SPlus tests whether bounded whitening and averaging capture much of that gain more stably. NorMuon and AdaMuon test the missing-variance hypothesis. Mousse and COSMOS test where expensive geometry should be concentrated.

\section{The emerging layerwise optimizer stack}

Matrix optimization is moving from one rule per model toward one rule per parameter class. This direction is visible even in methods that do not advertise it: Muon commonly uses AdamW for non-hidden parameters; SCALE reserves momentum for the output layer; symmetry-compatible designs derive distinct transforms for embeddings and routers.

\subsection{Equivariance as a routing principle}

Let a parameter transformation $T$ leave the represented network function unchanged after a corresponding transformation elsewhere in the model. An optimizer update $U$ is symmetry-compatible if
\begin{equation}
U(TW,T^{-T}G)=T\,U(W,G)
\end{equation}
for the relevant group action. Bi-orthogonal equivariance motivates polar updates for generic matrices. Embedding rows instead admit permutation symmetries tied to token identities; mixture routers admit expert permutations; gated MLP factors share coupled rescalings. A single coordinatewise Adam rule is not invariant to all of these actions, while a single two-sided spectral rule may impose the wrong invariance.

The symmetry-compatible optimizer derives one-sided spectral, row-norm, centered, and hybrid updates for these parameter classes \citep{lau2026symmetry}. Its experiments on dense and sparse models support the routing principle, but the broader research opportunity is larger: the architecture can expose parameter metadata, and the optimizer can dispatch a geometry accordingly.

\subsection{The cost of compositionality}

A layerwise stack introduces new hyperparameters and failure surfaces:
\begin{itemize}
  \item relative learning-rate scales between parameter classes;
  \item inconsistent weight-decay semantics;
  \item different state precisions and sharding layouts;
  \item update imbalance at residual branches;
  \item harder attribution when a combined method wins.
\end{itemize}
The natural engineering response is a common calibration layer. Update-RMS matching places each rule on a comparable scale. Token-based timescales align momentum across differently sharded batches. A routing table makes non-matrix fallbacks explicit. These should be treated as part of the algorithm specification.

\begin{table}[t]
\centering
\caption{Illustrative parameter-class routing for a compositional optimizer. The table states hypotheses to test, not a universal prescription.}
\label{tab:routing}
\small
\rowcolors{2}{soft}{white}
\begin{tabularx}{\textwidth}{P{0.20\textwidth}P{0.23\textwidth}Y Y}
\toprule
\textbf{Parameter class} & \textbf{Candidate geometry} & \textbf{Reason} & \textbf{Critical diagnostic} \\
\midrule
Attention/MLP hidden matrices & Muon, SOAP, SPlus, NorMuon & Dense two-axis structure; matrix kernels amortize well & Update RMS, singular-value spread, row-norm balance \\
Embedding tables & Row-aware Adam or one-sided rule & Sparse/non-uniform row access; token permutation symmetry & Frequency-conditioned update norm \\
Output head & Adam-like or dedicated momentum & High gradient variance and direct loss coupling & Logit-scale drift and row variance \\
Normalization scales/biases & AdamW or simple coordinate rule & One-dimensional parameters; matrix geometry is artificial & Relative update-to-weight ratio \\
MoE routers & Centered row/column-aware rule & Expert permutation and shared-shift structure & Expert-load stability and router entropy \\
\bottomrule
\end{tabularx}
\end{table}

\section{Memory is part of the algorithm}

For a parameter count $N$, full-precision Adam commonly stores two moment values per parameter in addition to weights, gradients, and often master weights. At scale, the question is not simply whether to precondition, but \emph{where information deserves persistent state}.

\subsection{Compress the statistic, not automatically the update}

\work{Adam-mini} preserves the first moment but replaces many elementwise second moments with blockwise statistics chosen to follow approximate Hessian structure \citep{zhang2024adammini}. It reports roughly 45--50\% lower optimizer-state memory than AdamW. \work{SlimAdam} makes compression conditional: a layerwise signal-to-noise criterion determines when second moments can be aggregated across dimensions, with up to 98\% second-moment savings reported in its experiments \citep{kalra2025slimadam}.

\work{APOLLO} uses random low-rank auxiliary state to estimate structured learning-rate scaling; its rank-one \work{Mini} variant reduces that state further \citep{zhu2024apollo}. \work{LDAdam} also maintains moments in a low-dimensional subspace, but explicitly transports state when the projection changes and uses generalized error feedback \citep{robert2024ldadam}. That machinery addresses a subtle failure mode: a moving basis can make a perfectly reasonable EMA semantically stale.

\work{GaLore2} improves the practicality of low-rank gradient projection, reducing projection-update and sharding bottlenecks and demonstrating 7B pretraining over long token horizons \citep{su2025galore2}. \work{Alice} and \work{RACS} begin from a structured Fisher approximation: Alice combines low-rank eigenspaces with Adam-like updates, while RACS uses row/column-scaled SGD with SGD-like memory \citep{gong2025structured}. Their study reports more than a twofold target-loss speedup for Alice over Adam in tested setups up to 1B parameters.

A crucial distinction runs through these methods:
\begin{center}
\fcolorbox{rulegray}{soft}{\parbox{0.88\textwidth}{
\textbf{Low-rank state is not the same as a low-rank update.}
One can compress the information used to choose scales while retaining a full-rank residual update. Conversely, a low-rank update can remain constrained even if its state estimate is accurate.
}}
\end{center}

\work{FRUGAL} makes the distinction explicit by assigning an advanced optimizer to a low-dimensional component while sending the full-rank residual through a state-free rule \citep{zmushko2024frugal}. \work{LoRA-Pre} reinterprets an EMA as online linear regression and factorizes the momentum itself \citep{wang2026lorapre}. Whether either is attractive depends on how quickly the relevant subspace changes.

\subsection{State-free and near-state-free alternatives}

\work{SWAN} uses instantaneous normalization and whitening without persistent moments \citep{ma2024swan}. The authors report SGD-level optimizer memory, roughly 50\% lower end-to-end memory, and about twofold token efficiency over Adam in 350M and 1.3B experiments. \work{SCALE} uses column-normalized SGD and keeps momentum only for the output layer; its 60M--1B results report Adam-like performance at 35--45\% of total training memory \citep{glentis2025scale}.

These methods are appealing when optimizer state determines the largest trainable model. Their caveat is equally important: instantaneous normalization has no temporal variance estimate, so batch size, gradient noise, and tensor shape become more exposed. A fixed-memory benchmark is more informative here than a table of bytes per parameter. Extra memory can be reinvested in batch size, sequence length, activation checkpointing, or model width, and each changes the optimization problem.

\section{Memory-efficient optimization as statistical compression}

Memory-efficient optimizers are sometimes compared by a single number: state bytes per parameter. That number is necessary but incomplete. The methods compress different statistical objects and therefore introduce different biases. A useful classification has four categories:
\begin{enumerate}
  \item \textbf{factorization}: replace an elementwise tensor with row, column, block, or Kronecker factors;
  \item \textbf{projection}: store adaptive state in a lower-dimensional subspace;
  \item \textbf{sparsification}: retain state for selected coordinates or parameter classes;
  \item \textbf{elimination}: use instantaneous normalization or plain SGD so no persistent state is required.
\end{enumerate}

Adafactor is an early example of factorizing a matrix second moment into row and column statistics \citep{shazeer2018adafactor}. Adam-mini uses a block partition informed by approximate Hessian structure rather than a rank-one factorization \citep{zhang2024adammini}. SlimAdam observes the state trajectory and compresses only dimensions whose signal-to-noise ratio makes aggregation safe \citep{kalra2025slimadam}. These three methods can have similar state counts on a particular model while encoding different assumptions.

\subsection{A concrete memory ledger}

Consider mixed-precision training with $N$ parameters. A simplified per-parameter ledger may include:
\begin{itemize}
  \item model parameters in BF16 or FP16: 2 bytes;
  \item gradients in BF16 or FP16: 2 bytes;
  \item FP32 master weights: 4 bytes;
  \item Adam first and second moments in FP32: 8 bytes.
\end{itemize}
Before activations and fragmentation, this is roughly 16 bytes per parameter. Sharding changes the per-device share, but not the aggregate volume. Removing one Adam moment, quantizing both, or eliminating master weights therefore changes different portions of the ledger.

The end-to-end benefit depends on the rest of the workload. If activations dominate, reducing optimizer state may not change the feasible batch. If parameters and state dominate, as in sparse mixture-of-experts models, eliminating master weights can be decisive. Papers should report both optimizer-only and total peak memory, with and without sharding.

\begin{table}[t]
\centering
\caption{What is compressed by representative memory-efficient methods.}
\label{tab:memory-compression}
\small
\rowcolors{2}{soft}{white}
\begin{tabularx}{\textwidth}{P{0.18\textwidth}P{0.22\textwidth}Y Y}
\toprule
\textbf{Method} & \textbf{Compressed object} & \textbf{Information retained} & \textbf{Main risk} \\
\midrule
Adam-mini & Second moment within Hessian-informed blocks & Full first moment; one scale per block & Partition may not match changing curvature \\
SlimAdam & Selected dimensions of second moments & Fine state where measured SNR is low & SNR estimate and decision overhead \\
APOLLO & Adaptive scaling in random low-rank coordinates & Structured learning-rate information & Projection may miss important directions \\
GaLore2 & Gradient and optimizer state in refreshed low-rank subspace & Full parameters; low-rank adaptive dynamics & SVD/refresh and subspace drift \\
LDAdam & Projected first/second moments & State transported between subspaces & Error-feedback stability and refresh cost \\
FRUGAL & Expensive optimizer on low-dimensional component & Full-rank residual via cheap rule & Balancing component and residual \\
SWAN & All persistent moments & Instantaneous normalization and whitening & Batch noise and shape sensitivity \\
SCALE & Most momentum and all second moments & Column normalization; momentum at output & Architecture-specific concentration of variance \\
\bottomrule
\end{tabularx}
\end{table}

\subsection{Projection changes coordinates over time}

Let $P_t\in\mathbb{R}^{d\times r}$ be an orthonormal basis with $r\ll d$. A projected optimizer forms
\begin{equation}
z_t=P_t^\top g_t
\end{equation}
and stores moments in $\mathbb{R}^r$. The full-space update may be reconstructed as $P_tu_t$. If $P_t$ changes, the old state is expressed in the wrong coordinates. Simply keeping the vector $m_{t-1}$ while replacing $P_{t-1}$ by $P_t$ changes its meaning.

LDAdam addresses this by transporting optimizer states between consecutive subspaces and applying generalized error feedback \citep{robert2024ldadam}. GaLore periodically recomputes a low-rank gradient basis, while GaLore2 reduces the SVD and distributed-integration bottlenecks and demonstrates much longer 7B pretraining \citep{zhao2024galore,su2025galore2}. The engineering advance is scientifically relevant: a projection that is refreshed too rarely becomes stale; one refreshed too often can spend its memory savings on decompositions and synchronization.

APOLLO avoids expensive data-dependent subspace estimation by using random projections to approximate structured learning-rate scaling \citep{zhu2024apollo}. The rank-one APOLLO-Mini result suggests that learning-rate information may be far more compressible than the gradient itself. This supports a general design rule:
\begin{quote}
\textbf{Use a compressed statistic to choose a full-space scale whenever possible; do not force the effective update to inherit the statistic's rank.}
\end{quote}

FRUGAL implements this rule by applying an advanced optimizer to a low-dimensional component and a stateless optimizer to the residual \citep{zmushko2024frugal}. RACS reaches a similar goal through row/column scaling, while Alice learns a low-rank eigenspace for richer adaptation \citep{gong2025structured}. The methods differ in how they divide expensive and cheap information, but all reject the idea that every direction deserves an independent second moment.

\subsection{When state should be allocated unevenly}

Optimizer state is most valuable where gradient statistics are heterogeneous and persistent. SCALE's finding that output-layer momentum carries disproportionate value is an extreme instance \citep{glentis2025scale}. SlimAdam's SNR rule gives a data-dependent version. Sparse SPAM retains momentum for selected coordinates. A layerwise stack can allocate:
\begin{itemize}
  \item full first and second moments to high-variance heads or embeddings;
  \item rowwise or blockwise state to hidden matrices;
  \item no state to stable normalization or residual parameters;
  \item low-rank matrix statistics only where measured anisotropy justifies them.
\end{itemize}

The unresolved question is whether this allocation can be predicted before training. Architecture, initialization, data mixture, and token frequency all affect compressibility. Online allocation is more robust but consumes monitoring and can create discontinuities when state is added or removed.

\subsection{Memory savings alter the statistical experiment}

Suppose optimizer A needs half the state of optimizer B. There are at least three fair comparisons:
\begin{enumerate}
  \item hold model and batch fixed, isolating the update rule;
  \item hold peak memory fixed and give A a larger batch or context;
  \item hold hardware and wall-clock fixed, letting each implementation choose its throughput-maximizing workload.
\end{enumerate}
The first is a scientific ablation; the second is a resource comparison; the third is a deployment comparison. They answer different questions. Reporting only the first can understate the value of compression, while reporting only the second can attribute a batch-size effect to the optimizer.

APOLLO reports a threefold throughput gain in a setting where its memory savings permit a fourfold larger batch \citep{zhu2024apollo}. Adam-mini reports substantial throughput improvement in a multi-GPU setup \citep{zhang2024adammini}. These are valuable end-to-end results, but the causal chain should remain explicit:
\begin{equation}
\text{smaller state}\rightarrow\text{larger feasible workload}
\rightarrow\text{higher device utilization}\rightarrow\text{wall-clock gain}.
\end{equation}
The optimizer may not be faster at an equal batch; it enables a faster training configuration.

\subsection{A fixed-memory evaluation frontier}

The most informative output is a Pareto frontier with at least three axes: validation loss, wall-clock time, and peak memory. A fourth axis can report robustness, such as the fraction of a learning-rate sweep that remains stable. For each point, the paper should state:
\begin{itemize}
  \item rank, block size, or state-allocation policy;
  \item refresh interval and factorization cost;
  \item whether projection constrains the final update;
  \item total and optimizer-only memory;
  \item batch and sequence length enabled by the configuration.
\end{itemize}

Without this information, a low-memory optimizer can look either better or worse merely because the saved memory was or was not reinvested. The correct scientific object is not bytes per parameter in isolation, but the quality-cost frontier of the complete training configuration.

\section{Time: schedules, averaging, and unknown horizons}

The learning-rate schedule is often treated as external to the optimizer. That is increasingly untenable. Momentum defines a timescale, weight decay accumulates over time, averaging selects among iterates, and the final decay phase can change both validation loss and downstream adaptability.

\subsection{Schedule-free does not mean time-free}

\work{The Road Less Scheduled} constructs schedule-free variants by maintaining fast and averaged iterates and combining momentum with online-to-batch conversion \citep{defazio2024schedulefree}. Hyperparameters need not encode the final training horizon. This does \emph{not} remove temporal policy; it internalizes it in interpolation and averaging.

Warmup-stable-decay (WSD) and reusable cooldowns attack the same horizon problem from a checkpointing perspective \citep{wen2024wsd,hagele2024cooldown}. A stable high-rate trunk can continue without a known endpoint, and a short decay branch exposes a strong checkpoint when needed. \work{Weight-space model merging} shows that merging checkpoints along a trajectory can emulate some effects of learning-rate decay, with merge duration acting as an important control variable \citep{tian2025wsm}.

The matrix-aware versions are now arriving. \work{SF-NorMuon} combines schedule-free training with neuron-normalized spectral updates and reports competitive results for 125M and 772M models across $1\times$--$8\times$ horizons \citep{apte2026sfnormuon}. \work{ScheduleFree+} reports long-horizon language-model results and a 31\% advantage over a WSD baseline at a very high tokens-per-parameter ratio \citep{defazio2026schedulefreeplus}. Both are promising but still emerging. Notably, SF-NorMuon finds that applying weight decay at the fast iterate is essential, a reminder that averaging, decay, and geometry cannot always be composed naively.

\work{Weight-space optimization without decay} complicates the endpoint story further: in its 1B and 8B studies, a checkpoint with worse pretraining loss can adapt better during downstream supervised fine-tuning \citep{yano2026wso}. Decay can improve the pretraining endpoint while reducing later plasticity. Therefore best final loss is not always the correct training objective.

\subsection{Batch size changes the optimizer's clock}

An EMA decay is normally specified per update, but examples arrive per token. If a reference run uses batch $B$ and decay $\beta$, then a run with batch $B'$ preserves approximately the same half-life in processed examples by using
\begin{equation}
\beta'=\beta^{B'/B}.
\label{eq:batchbeta}
\end{equation}
Small-batch experiments show that applying this conversion can make momentum SGD surprisingly competitive and that gradient accumulation may waste optimization opportunities even when it improves device utilization \citep{marek2025smallbatch}. Before concluding that an optimizer fails at a new batch size, one should first ask whether its memory was measured in steps or in tokens.

\section{Horizon management in detail}

A learning-rate curve is a policy over training time. It decides how long the process explores, when it reveals the progress accumulated during a high-rate phase, and whether a useful checkpoint exists before the planned endpoint. Three families now address this policy: explicit decay, stable trunks with cooldowns, and schedule-free averaging.

\subsection{Cosine decay and the hidden endpoint}

Cosine decay specifies a horizon $T$:
\begin{equation}
\eta_t=\eta_{\min}+
\frac{1}{2}(\eta_{\max}-\eta_{\min})
\left(1+\cos\frac{\pi t}{T}\right).
\end{equation}
The endpoint enters every step. Extending a run changes the schedule that would have been preferred from the beginning. A checkpoint at $0.5T$ was trained under a rate selected for a future decay and is not equivalent to the midpoint of a run planned for $2T$.

This path dependence complicates scaling studies and data additions. It also helps explain optimizer-ranking flips. A method that tolerates a high learning rate can lead during the stable portion, while a variance-adaptive method can benefit more from late decay. Comparing before the intended endpoint can favor the wrong mechanism \citep{wen2025fantastic}.

\subsection{Stable trunks and cooldown branches}

Warmup-stable-decay retains a high or constant rate for a reusable main trajectory, then starts a decay branch when a checkpoint is needed \citep{wen2024wsd}. The river-valley interpretation separates motion along a flat valley from oscillation across steep walls: the stable rate makes longitudinal progress while the cooldown suppresses transverse oscillation.

Reusable cooldown studies show that constant-rate training followed by a short decay can obey predictable scaling behavior and make one long trajectory support multiple endpoints \citep{hagele2024cooldown}. The practical benefit is option value. A project can continue the trunk when more data or compute arrives without retroactively changing all previous rates.

The trunk is not fully horizon-free if its optimal peak rate depends on the intended token budget. Power Scheduler models this dependence by relating learning rate to processed tokens and batch size \citep{shen2024power}. In its empirical scaling rule,
\begin{equation}
\eta_{\mathrm{opt}}\approx B\,aT^b,
\end{equation}
with $b$ near a negative half in the studied regime. The resulting online schedule depends on tokens already processed rather than a fixed final horizon. The important conceptual step is to express time in tokens and to model transfer across batch and model scale.

\subsection{Schedule-free interpolation}

Schedule-free optimization maintains multiple iterates. A fast point follows gradient information; an averaged point accumulates stability; an interpolated point is used to evaluate gradients \citep{defazio2024schedulefree}. Although implementations differ, the generic pattern is
\begin{align}
y_t &= (1-\gamma_t)z_t+\gamma_t x_t,\\
z_{t+1} &= z_t-\eta\,g(y_t),\\
x_{t+1} &= (1-c_t)x_t+c_t z_{t+1}.
\end{align}
The coefficients induce effective decay and averaging without encoding the final step. The method is schedule-free in the sense of horizon independence, not in the sense of constant dynamics.

This semantic distinction matters when composing weight decay. Applying decay to the averaged iterate, fast iterate, or gradient-evaluation point produces different trajectories. SF-NorMuon identifies decay at the fast iterate as essential in its long-horizon spectral construction \citep{apte2026sfnormuon}. A paper that reports only a decay coefficient without its location has not fully specified the algorithm.

ScheduleFree+ adds scale and batch corrections aimed at long language-model runs \citep{defazio2026schedulefreeplus}. Its largest claimed benefit appears at an extreme tokens-per-parameter ratio, which is exactly where a horizon-independent averaging policy should be most differentiated from a short cosine run. Independent large-scale replication remains important.

\subsection{Checkpoint merging as temporal optimization}

Weight-space model merging offers another view of decay. Averaging or merging checkpoints from a trajectory can suppress high-frequency parameter noise and approximate the endpoint produced by a cooldown \citep{tian2025wsm}. The merge window becomes a temporal hyperparameter: a short window behaves like local averaging; a long window can mix functionally distant models and blur useful progress.

This connection unifies three practices often discussed separately:
\begin{itemize}
  \item stochastic weight averaging;
  \item schedule-free online averaging;
  \item post-hoc checkpoint merging.
\end{itemize}
All choose a temporal kernel over iterates. The differences are whether the kernel affects subsequent gradients, whether it can be updated online, and how much state or checkpoint storage is required.

\subsection{Dual averaging and regularization schedules}

SODA interprets several modern optimizers through optimistic dual averaging and proposes a wrapper with a theoretically motivated $1/k$ decay rule \citep{pethick2026soda}. In this view, Muon, Lion, AdEMAMix, and NAdam differ in the geometry and optimism used to map accumulated gradients into a primal update. The wrapper aims to reduce weight-decay tuning rather than replace the base optimizer.

This is useful because weight decay has at least three roles:
\begin{enumerate}
  \item regularization toward small weights;
  \item control of equilibrium weight norms;
  \item indirect control of angular learning rate.
\end{enumerate}
A $1/k$ schedule, constant decoupled decay, norm projection, and cautious decay implement different policies even when they produce similar final norms. Hyperball and Muown make the angular interpretation explicit, while no-decay weight-space studies ask whether preserving plasticity can matter more than minimizing pretraining loss \citep{wen2026hyperball,lion2026muown,yano2026wso}.

\subsection{Batch, accumulation, and the unit of time}

Equation~\eqref{eq:batchbeta} preserves an EMA half-life across batch sizes, but learning rate and noise scale also change. Gradient accumulation is especially subtle. Accumulating $K$ microbatches and taking one optimizer step is not equivalent to taking $K$ smaller steps:
\begin{itemize}
  \item the parameters remain fixed across the $K$ microbatches;
  \item clipping and normalization act on the accumulated gradient;
  \item momentum and second moments are updated once rather than $K$ times;
  \item matrix factors or probes are computed from a different statistic.
\end{itemize}

For wall-clock efficiency, accumulation may be unavoidable. For optimizer comparison, the effective update frequency should be reported in both steps and tokens. If a method benefits from frequent state updates, a large accumulated batch can erase its advantage.

\begin{table}[t]
\centering
\caption{Horizon strategies and the question each one answers.}
\label{tab:horizon}
\small
\rowcolors{2}{soft}{white}
\begin{tabularx}{\textwidth}{P{0.18\textwidth}P{0.23\textwidth}Y Y}
\toprule
\textbf{Strategy} & \textbf{Endpoint assumption} & \textbf{Strength} & \textbf{Primary caveat} \\
\midrule
Cosine decay & Fixed $T$ known in advance & Strong mature baseline and smooth endpoint & Earlier iterates depend on planned endpoint \\
WSD/cooldown & Stable trunk, endpoint chosen by branch & Reusable trajectory and multiple budgets & Peak rate and cooldown still need calibration \\
Power schedule & Depends on processed tokens, not final $T$ & Transfers time policy across token and batch scales & Empirical power law may be regime-specific \\
Schedule-free & Unknown horizon & Anytime averaged iterate with constant main scale & Averaging and decay placement are algorithmic choices \\
Checkpoint merging & Post-hoc or periodic & Converts stored trajectory into implicit decay & Merge window, storage, and function-space mismatch \\
\bottomrule
\end{tabularx}
\end{table}

\section{Low precision: state semantics matter more than bit count}

Quantizing weights and activations while keeping optimizer states in high precision leaves a large memory opportunity untouched. Yet moments are not ordinary activations: they are recursively updated state, and small rounding errors can alter their effective timescale.

\work{COAT} stores optimizer state and selected training tensors in FP8 using dynamic range expansion and mixed-granularity scaling \citep{xi2024coat}. It reports up to $1.54\times$ memory reduction and $1.43\times$ speedup over a BF16 baseline. \work{SOLO} targets 2--3 bit state using logarithmic unsigned quantization and precision-specific momentum, reporting roughly 45GB of savings for a 7B model \citep{xu2025solo}.

Muon has an interesting advantage in this regime. A study of 8-bit Muon finds blockwise dynamic or linear quantization sufficient to match full-precision Muon at 1.6B scale, while reducing optimizer state by 74\% relative to full-precision Muon and 86\% relative to FP32 AdamW \citep{gupta2025quantmuon}. The explanation is structural: Adam's second moment appears in a denominator, where small or poorly quantized values can cause large effective steps. Momentum used only before a polar transform is more tolerant of scale error.

The less obvious problem is \emph{quantization staleness}. When an EMA update is smaller than a quantization cell, rounding can return the stored value unchanged; the moment silently stops responding. Periodic state resets can restore responsiveness \citep{topollai2026statequant}. \work{ECO} removes full-precision master weights and feeds weight-quantization error back into momentum, extending low-precision reasoning beyond moments \citep{nikdan2026eco}.

Non-smooth transitions create another failure mode. Quantizers, hard routers, and structured sparsity can make nearby parameter perturbations produce sharply different loss responses. Singularity-aware Adam uses randomized probes to detect this local disagreement and reduces the update only while the instability persists \citep{xu2026singularity}. The idea is complementary to state resets: resets remove contaminated history, whereas probing tries to recognize when the current neighborhood itself is unreliable. The additional objective evaluations make it most plausible when instability is costly enough to justify an explicit diagnostic.

The design principle is broader than any format:
\begin{quote}
\textbf{Quantize according to the state's topology and recurrence, not only its histogram.}
\end{quote}
An unsigned logarithmic representation suits a nonnegative second moment; a blockwise signed representation suits momentum; an error-feedback channel suits a repeatedly quantized parameter. The phrase eight-bit optimizer is therefore not a complete algorithm description.

\section{Quantized optimizer dynamics}

Quantization maps a real state $s$ to a finite codebook:
\begin{equation}
\widehat s=Q(s;\alpha,\mathcal{C}),
\end{equation}
where $\alpha$ is a scale and $\mathcal{C}$ is a set of representable codes. For an activation used once, quantization error disappears after the layer computation. For an optimizer state, the quantized value is fed back into the next update:
\begin{equation}
\widehat s_t=Q\!\left(\beta\widehat s_{t-1}+(1-\beta)x_t\right).
\label{eq:quantized-ema}
\end{equation}
The error is recursive. It can change the effective decay, create absorbing codebook cells, and interact with resets.

\subsection{Blockwise dynamic quantization as the baseline}

Early 8-bit optimizers combine blockwise scaling, nonlinear dynamic codebooks, and special handling of unstable embeddings \citep{dettmers2021eightbit}. Blockwise scaling isolates outliers: one extreme coordinate changes the resolution of its block rather than an entire tensor. Dynamic codebooks devote more levels to the small values common in moment tensors.

This remains a strong engineering baseline. A new low-bit optimizer should compare against a tuned blockwise implementation, not naive per-tensor linear quantization. It should also distinguish storage precision from compute precision: many kernels dequantize a state in registers, perform the arithmetic at higher precision, and requantize only for memory.

\subsection{Why first and second moments need different codebooks}

Momentum is signed and often roughly symmetric, while Adam's second moment is nonnegative and heavy-tailed. Their downstream use differs:
\begin{equation}
u_t=\frac{m_t}{\sqrt{v_t}+\epsilon}.
\end{equation}
An additive error in $m_t$ perturbs the numerator. An error that underestimates a small $v_t$ can greatly enlarge the effective step. This asymmetry explains why a single linear quantizer is rarely optimal for both states.

SOLO identifies two low-bit pathologies: signal swamping in unsigned states and direction corruption in signed states \citep{xu2025solo}. It uses logarithmic unsigned quantization for the second moment and adjusts momentum according to precision. This is a good example of algorithm--representation co-design: the optimizer hyperparameter is changed because the storage recurrence changes.

COAT instead targets a broader FP8 training stack, using dynamic range expansion for state and mixed-granularity quantization for activations \citep{xi2024coat}. Its speed benefit depends on kernels for more than the optimizer. The comparison is therefore end-to-end rather than a pure update-rule ablation.

\subsection{State staleness}

Let $\Delta_t=(1-\beta)(x_t-\widehat s_{t-1})$ be the unquantized increment in Eq.~\eqref{eq:quantized-ema}. If $|\Delta_t|$ is smaller than half the local quantization interval, then
\begin{equation}
Q(\widehat s_{t-1}+\Delta_t)=\widehat s_{t-1}.
\end{equation}
The nominal update occurs in floating point but the stored state does not move. Repeated stalling makes the effective half-life much longer than $(1-\beta)^{-1}$.

The state-staleness analysis models this probability and explains why resets can restore responsiveness \citep{topollai2026statequant}. A reset moves the state away from an absorbing local cell and temporarily increases increments. But resetting too often destroys useful history. A theory-guided period should depend on bit width, scale policy, $\beta$, and the distribution of observations.

This mechanism connects low precision to SPAM. In high precision, resets remove contamination by rare spikes. In low precision, they can also remove accumulated staleness. Stable-SPAM combines state reset with norm tracking and clipping for 4-bit training \citep{huang2025stablespam}. The same operation has two different justifications, which should be diagnosed separately.

\subsection{Why Muon is unusually quantization tolerant}

Muon stores momentum but subsequently maps it through an approximate polar transform. If quantization mostly perturbs singular values while preserving dominant singular subspaces, the polar map removes much of the scale error. Adam places the quantized second moment in a denominator, where scale error is directly converted to step-size error.

The 8-bit Muon study reports parity with full-precision Muon up to 2.7B parameters and as much as 62\% reduction in optimizer-state footprint in its latest experiments \citep{gupta2025quantmuon}. It finds that simple linear blockwise quantization can work for Muon, whereas AdamW more often requires a dynamic codebook. This is not evidence that every spectral optimizer is quantization invariant. Perturbations that rotate singular subspaces or interact with finite Newton--Schulz iterations can still matter.

\subsection{Eliminating master weights}

Mixed-precision training commonly applies an update to an FP32 master copy and then casts to the training format. If the update is applied directly to a quantized weight,
\begin{equation}
\widehat W_{t+1}=Q_W(\widehat W_t+\Delta W_t),
\end{equation}
small updates can disappear and the quantization residual accumulates.

ECO injects this residual into momentum as error feedback \citep{nikdan2026eco}. Conceptually,
\begin{align}
\widetilde W_{t+1} &= \widehat W_t+\Delta W_t+e_t,\\
\widehat W_{t+1} &= Q_W(\widetilde W_{t+1}),\\
e_{t+1} &= \widetilde W_{t+1}-\widehat W_{t+1},
\end{align}
with the error represented through the optimizer dynamics without a full additional buffer. The method targets a different memory term from state quantization. Combining low-bit moments and no master weights can therefore compound savings, but also couples two recursive error channels.

\subsection{Dynamic precision allocation}

Fixed bit width ignores the fact that sensitivity varies by layer, state, and training phase. STQuant allocates precision spatio-temporally, selecting influential factors and updating decisions with lower search cost \citep{liu2026stquant}. Its reported average width near five bits and large state-memory reduction suggest that uniform 8-bit storage leaves efficiency on the table.

Dynamic allocation introduces a control problem. A precision transition changes the quantization lattice and can itself perturb state. The controller needs a stable sensitivity signal, a transition cost, and a budget constraint. Useful evaluations should report the precision path over time, not only its average.

\subsection{Optimizer choice can change later quantizability}

Low-precision optimization is not limited to storing the training state. The optimizer can change the activation outliers of the final model. Outlier-Safe Pre-Training combines Muon, a single-scale normalization design, and an embedding projection to prevent extreme activation channels \citep{park2025outliersafe}. A 1.4B model trained for one trillion tokens shows much stronger 4-bit post-training behavior than an Adam-trained comparison in that study.

The causal contribution of the optimizer must be separated from the architecture changes, but the result broadens the evaluation target. A training optimizer can be judged by:
\begin{itemize}
  \item pretraining loss and time;
  \item state memory during training;
  \item stability under quantization-aware training;
  \item outlier statistics and post-training quantizability.
\end{itemize}
An optimizer that is slightly worse in BF16 loss but produces a substantially easier-to-quantize model may be preferable for deployment.

\begin{table}[t]
\centering
\caption{Failure modes specific to low-precision optimizer state.}
\label{tab:quant-failures}
\small
\rowcolors{2}{soft}{white}
\begin{tabularx}{\textwidth}{P{0.19\textwidth}Y Y Y}
\toprule
\textbf{Failure} & \textbf{Mechanism} & \textbf{Diagnostic} & \textbf{Mitigation} \\
\midrule
Range saturation & Outlier exceeds block scale & Saturated-code fraction by layer & Smaller blocks, dynamic range expansion \\
State staleness & EMA increment rounds to same code & Fraction of unchanged state entries & Log codebook, scale update, reset \\
Denominator error & Small $v_t$ is under-resolved & Distribution of effective learning rates & Unsigned log quantization, floor, higher precision \\
Direction corruption & Signed momentum crosses code boundary & Cosine between quantized/full update & Precision-specific momentum, error feedback \\
Master-weight loss & Parameter update vanishes after requantization & Accumulated weight residual & Master copy or ECO-style feedback \\
\bottomrule
\end{tabularx}
\end{table}

\section{A synthesis: five claims that survive the leaderboard}

\subsection{Matrix-aware optimization is real, but second order is too coarse}

Muon's polar normalization, Shampoo's Kronecker whitening, SOAP's rotated Adam, and row/column-adaptive hybrids solve different problems. Their gains across controlled and scaled studies are too consistent to dismiss. But calling all of them second order encourages incorrect transfer of intuition and hyperparameters. The relevant questions are which statistics are historical, which norm is controlled, and which information is discarded.

\subsection{Variance adaptation keeps returning}

Pure spectral normalization is strong, especially early in training and at shorter data ratios. Longer horizons often reward variance-aware methods, whether diagonal, rotated, rowwise, or Kronecker-factored \citep{wen2025fantastic,frans2025matters}. AdaMuon, NorMuon, COSMOS, and Mousse can all be read as attempts to recover variance or curvature information at different granularities.

\subsection{The best geometry is parameter-class dependent}

Most Muon implementations already exempt vectors, biases, normalization scales, and often embeddings, sending them to AdamW. A symmetry-compatible optimizer makes the principle explicit: assign an update rule based on the symmetry and role of each parameter block, including embeddings, output heads, gated MLPs, and routers \citep{lau2026symmetry}. One optimizer for the model may eventually sound as crude as one initialization scale for every layer.

\subsection{Memory and systems can reverse an algorithmic result}

Matrix inverse roots, basis updates, all-gathers, state sharding, and quantization kernels determine wall-clock performance. An exact distributed SOAP that hides communication is scientifically different from an approximation that changes its preconditioner \citep{khona2026soapmuon}. Likewise, a smaller optimizer state can enable a batch or sequence length that changes the gradient distribution. Reported convergence without system context is incomplete.

\subsection{There is no universal twofold replacement for AdamW}

The strongest current conclusion is conditional. A well-tuned matrix-aware method can reduce tokens to target loss, sometimes substantially. A temporal method such as AdEMAMix or MARS can win in a different large-batch, long-horizon regime. AdamW can remain preferable when robustness, downstream compatibility, implementation maturity, or switching cost dominates. The honest output of the literature is a decision procedure, not a crown.

\section{Why optimizer leaderboards disagree}

The optimizer literature contains apparent contradictions:
\begin{itemize}
  \item Muon leads some short-horizon comparisons, while SOAP or Kronecker methods lead at longer data ratios.
  \item AdEMAMix and MARS lead one large-batch study, while matrix methods dominate another.
  \item a method reports a large speedup over AdamW in its own paper, but a controlled benchmark recovers a smaller gap.
  \item an intermediate checkpoint favors one optimizer and the final checkpoint reverses the order.
\end{itemize}
These observations are compatible once the protocol is treated as part of the result.

\subsection{Scale and data ratio are separate axes}

Model scale changes width, depth, matrix shapes, communication, and the relative cost of optimizer kernels. Data-to-parameter ratio changes how long the optimizer remains in early, middle, and terminal regimes. A study spanning 0.1B--1.2B parameters and $1\times$--$8\times$ a standard compute-optimal data ratio finds that matrix-optimizer gains shrink with model scale while rankings also change with training duration \citep{wen2025fantastic}.

This means that a single scaling curve is insufficient. At minimum, evaluation needs a grid:
\begin{equation}
\{\text{model scale}\}\times\{\text{tokens per parameter}\}.
\end{equation}
A method can have a favorable exponent in one dimension and unfavorable behavior in the other. Short proxy runs can then be misleading even when the proxy model is architecturally faithful.

\subsection{Batch size changes both noise and systems}

The comprehensive optimizer benchmark varies model size, batch, and horizon and finds different leading families across settings \citep{semenov2025benchmark}. Batch size changes:
\begin{itemize}
  \item stochastic gradient variance;
  \item the token half-life of momentum;
  \item the number of optimizer steps at a fixed token budget;
  \item hardware utilization and communication frequency.
\end{itemize}
A matrix preconditioner estimated from a large, low-noise batch may behave differently from the same code on a small batch. AdEMAMix's slow state sees a different number of updates. A schedule specified in steps changes in token time. Every one of these can alter rankings.

\subsection{Hyperparameter transfer can be unfair in both directions}

Blindly copying AdamW's learning rate to a matrix optimizer can under-scale or over-scale the update. But giving every method an unconstrained search can also be misleading if the tuning budgets differ. A fair study should distinguish:
\begin{enumerate}
  \item \textbf{native tuning}: best achievable result with a comparable search budget;
  \item \textbf{transfer tuning}: performance when parameters are transferred by a published scaling rule;
  \item \textbf{default robustness}: volume of the stable, near-optimal region.
\end{enumerate}

Update-RMS matching is a useful calibration, not a substitute for tuning. If an Adam step has RMS $r_A$ and a Muon step has RMS $r_M$, scale the initial learning rate so their early RMS values match. The optimizers can still diverge later because their norm dynamics differ.

Weight decay must be tuned jointly with learning rate. In scale-invariant networks, decay changes weight norms and hence angular progress. Matrix optimizers can appear fragile to decay because their update norm is more tightly controlled; Hyperball and Muown are direct responses to this interaction \citep{wen2026hyperball,lion2026muown}.

\subsection{The endpoint metric can hide practical value}

Final validation loss is important but not sufficient. A method may:
\begin{itemize}
  \item reach a useful target earlier but finish at the same loss;
  \item have better final pretraining loss but worse downstream plasticity;
  \item be token-efficient but wall-clock slower;
  \item be slightly worse per token but enable a much larger fixed-memory model;
  \item reduce loss spikes and failed runs without improving the median successful run.
\end{itemize}

The no-decay weight-space study suggests that pretraining endpoint and later supervised adaptation can diverge \citep{yano2026wso}. Stable-SPAM emphasizes failure prevention. Low-memory methods emphasize feasible workload. A single scalar cannot rank all of these objectives.

\begin{table}[t]
\centering
\caption{Protocol variables that can reverse an optimizer comparison.}
\label{tab:ranking-flips}
\small
\rowcolors{2}{soft}{white}
\begin{tabularx}{\textwidth}{P{0.19\textwidth}Y Y}
\toprule
\textbf{Variable} & \textbf{Mechanism of reversal} & \textbf{Minimum reporting requirement} \\
\midrule
Training horizon & Early spectral gain versus late variance adaptation; decay exposes progress & Multiple endpoints including planned final token budget \\
Model scale & Kernel-to-model cost and optimizer advantage scale differently & At least two scales; fit uncertainty if claiming a trend \\
Batch size & Changes noise, momentum half-life, step count, and utilization & Global tokens per step, microbatch, accumulation, $\beta$ conversion \\
Schedule & Different methods tolerate or prefer different rate paths & Method-specific schedule sweep or justified shared policy \\
Weight decay & Controls norms and angular rate, not only regularization & Joint learning-rate/decay sweep and weight-norm traces \\
Parameter routing & Non-matrix fallback can dominate embeddings and heads & Exact parameter-class map and relative update RMS \\
Precision & Quantization error affects denominators, factors, and polar maps differently & State/compute precision for every operation \\
Systems implementation & Communication and factorization cost alter wall-clock order & Step-time breakdown and exposed/overlapped communication \\
\bottomrule
\end{tabularx}
\end{table}

\subsection{Reading paper-specific speedup claims}

A claim such as 1.4 times faster should be decomposed into:
\begin{equation}
\text{speedup}
=
\frac{\text{baseline resource to target}}
{\text{method resource to target}}.
\end{equation}
The resource may be tokens, steps, FLOPs, or seconds. The target may be final loss, a chosen intermediate loss, or downstream score. The baseline may use transferred rather than tuned hyperparameters. None of these choices invalidates the claim, but all determine its scope.

For example, the scalable Muon study fits scaling laws and reports a large FLOP reduction in its own protocol \citep{liu2025muonscalable}. The controlled cross-scale study finds smaller gains at its largest model \citep{wen2025fantastic}. The large-scale systems study again finds both Muon and SOAP better than AdamW under update-RMS matching and very large batches \citep{khona2026soapmuon}. The correct synthesis is not an average percentage. It is that matrix geometry remains beneficial across diverse regimes, while the magnitude is strongly protocol-dependent.

\subsection{Evidence tiers should apply to claims, not papers}

A paper can contain scaled evidence for stability, controlled evidence for one component, and emerging evidence for generalization. Labeling the entire paper strong or weak loses this granularity. This survey uses:
\begin{itemize}
  \item \textbf{scaled claim}: demonstrated at high parameter/token/system scale;
  \item \textbf{controlled claim}: isolated through matched sweeps or ablations;
  \item \textbf{theoretical claim}: proved under stated assumptions;
  \item \textbf{emerging claim}: recent, narrow, or awaiting independent replication.
\end{itemize}

The label attaches to the sentence. Muon has scaled evidence that polar-style updates can work in large language models, controlled evidence for some scaling choices, and still-emerging evidence for many recent variants. S-Adam has a developed stability argument for non-smooth objectives and emerging large-model evidence. STQuant has a concrete systems result but limited architecture coverage. This claim-level discipline is more informative than ranking publication status.

\subsection{A reproducible benchmark matrix}

A compact but credible future benchmark could use:
\begin{enumerate}
  \item two model scales separated by at least fourfold parameters;
  \item two data ratios, one short and one long;
  \item two batch regimes with token-time-corrected momentum;
  \item cosine and an anytime schedule;
  \item full-precision and one low-precision state configuration;
  \item a fixed-token, fixed-FLOP, and fixed-memory view.
\end{enumerate}
Not every optimizer needs a full factorial search. A staged design can first tune each family on a central setting, then test transfer. The important requirement is to predeclare which comparisons isolate mechanisms and which measure complete recipes.

\begin{table}[t]
\centering
\caption{A minimal family-level benchmark rather than a name-level tournament.}
\label{tab:benchmark-families}
\small
\rowcolors{2}{soft}{white}
\begin{tabularx}{\textwidth}{P{0.20\textwidth}P{0.23\textwidth}Y Y}
\toprule
\textbf{Family} & \textbf{Representative baseline} & \textbf{Mechanism isolated} & \textbf{Required control} \\
\midrule
Coordinate adaptive & AdamW / ADOPT & Elementwise variance and update ordering & $\beta_2$, decay, denominator precision \\
Multi-timescale & AdEMAMix / MARS & Longer memory or variance-reduced temporal estimator & Token half-life and batch size \\
Spectral & Muon / Scion & Matrix norm and polar direction & Shape scaling and update RMS \\
Whitening & SOAP / SPlus & Historical matrix basis and variance adaptation & Factor refresh, state cost, eigensolver \\
Hybrid matrix & NorMuon / COSMOS & Granularity and placement of adaptation & Same polar implementation and fallback policy \\
Low-memory & Adam-mini / GaLore2 / SWAN & State support, projection, or elimination & Fixed batch and fixed-memory views \\
Low-precision & 8-bit AdamW / 8-bit Muon / SOLO & Quantized state recurrence & Same codebook budget and kernel accounting \\
\bottomrule
\end{tabularx}
\end{table}

\section{A practical decision guide}

\begin{table}[t]
\centering
\caption{Reasonable starting points by constraint. These are experiment priorities, not universal defaults.}
\label{tab:guide}
\small
\rowcolors{2}{soft}{white}
\begin{tabularx}{\textwidth}{P{0.20\textwidth}P{0.25\textwidth}Y}
\toprule
\textbf{Primary constraint} & \textbf{Start with} & \textbf{What to verify first} \\
\midrule
Low switching risk & Tuned AdamW; then Cautious or AdEMAMix &
Retune learning rate, decay, and momentum timescale; measure target-loss time, not only final loss \\
Token efficiency on matrix-heavy models & Muon as the inexpensive matrix baseline; SOAP/SPlus or a Kronecker method for longer horizons &
Update-RMS alignment, matrix eligibility rules, shape scaling, and non-matrix parameter treatment \\
Optimizer-state memory & Adam-mini or SlimAdam for Adam-like semantics; APOLLO/RACS/SWAN for aggressive reduction &
Fixed-memory throughput, basis-refresh cost, full-rank residual behavior, and accuracy at the enlarged batch \\
Unknown stopping time & Schedule-free AdamW or a WSD/cooldown trunk; SF-NorMuon as an emerging matrix option &
Anytime checkpoint quality, decay/averaging placement, and later fine-tuning plasticity \\
Low-precision state & A tested dynamic 8-bit AdamW implementation or 8-bit Muon &
State staleness, small-denominator behavior, loss-spike recovery, and end-to-end kernel speed \\
Small batch & Existing optimizer with token-time rescaling of $\beta$ &
Use Eq.~\eqref{eq:batchbeta}; compare true microbatch updates against gradient accumulation \\
\bottomrule
\end{tabularx}
\end{table}

A conservative experiment sequence is:

\begin{enumerate}
  \item \textbf{Establish a reproducible AdamW frontier.} Sweep learning rate and weight decay jointly; log update RMS by parameter class; keep data order and precision fixed.
  \item \textbf{Change one optimizer axis.} If testing Muon, keep the schedule conventional. If testing schedule-free training, start with AdamW geometry. If quantizing state, first match the full-precision algorithm.
  \item \textbf{Add the interaction only after both components work.} A schedule-free matrix optimizer or low-bit SOAP changes multiple axes and is otherwise hard to diagnose.
  \item \textbf{Choose the budget metric before the winner.} Fixed tokens answers a sample-efficiency question; fixed FLOPs includes matrix operations; time-to-target includes kernels and communication; fixed HBM includes the opportunity to enlarge the workload.
\end{enumerate}

\section{A minimum credible evaluation protocol}

Because optimizer rankings are protocol-sensitive, a credible comparison should report more than a final loss:

\begin{itemize}
  \item same initialization distribution, data mixture, token order, tokenizer, precision, and model parameterization;
  \item optimizer-specific sweeps for learning rate, momentum/decay constants, weight decay, and schedule---defaults should not be transplanted blindly;
  \item the policy for embeddings, output heads, normalization parameters, biases, and tensors that are not two-dimensional;
  \item update RMS and weight RMS by parameter class, especially when comparing Adam-like and spectral methods;
  \item loss at fixed tokens and FLOPs, time to a prespecified target, peak memory, optimizer-step time, and communication volume;
  \item multiple horizons, including the actual endpoint, because rankings can flip late;
  \item several seeds where the expected gap is small, plus instability counts and failed hyperparameter regions;
  \item downstream adaptation, retention, or quantization quality when those are the real deployment objectives.
\end{itemize}

The protocol should also state how much tuning compute each method received. An optimizer that wins only after a much larger sweep may still be worthwhile for a trillion-token run, but the search cost is part of the adoption decision.

\section{Open problems for the next optimizer cycle}

\textbf{A theory of hybrid geometry.}
We have useful interpretations of diagonal adaptivity, Kronecker preconditioning, and spectral steepest descent, but less guidance for choosing the granularity at which they should be combined. Why should variance be elementwise, rowwise, subspace-wise, or singular-direction-wise?

\textbf{Scale laws for optimizer advantage.}
The observed gap between matrix methods and Adam often shrinks with model size yet changes with data-to-parameter ratio \citep{wen2025fantastic}. We need scaling laws that include optimizer state, batch, horizon, and tuning budget, not only parameter and token counts.

\textbf{Optimizer-aware parameterization.}
Update-RMS matching is a repair for mismatched conventions. A cleaner approach would co-design initialization, residual scaling, normalization, and optimizer geometry so learning-rate transfer is a property of the parameterization.

\textbf{State under changing coordinates.}
Low-rank projections, eigenspaces, sharding layouts, and quantizers all create coordinate systems that change over time. State transport and error feedback should become first-class primitives rather than paper-specific patches.

\textbf{Anytime training with downstream objectives.}
Schedule-free and cooldown methods optimize access to checkpoints, while no-decay results suggest that pretraining loss and later plasticity can diverge \citep{yano2026wso}. The right objective may be a path of adaptable models rather than one endpoint.

\textbf{Reproducible systems evidence.}
The field needs public, exact, distributed implementations with matched kernels and communication. Otherwise algorithmic and engineering gains remain entangled in ways that cannot be audited.

\section{Conclusion}

The optimizer landscape of 2025--2026 is best understood as a move from scalar update rules to \emph{composed training policies}. Temporal estimation decides what history is trustworthy. Matrix geometry decides how a layer should move. Horizon policy decides when progress becomes visible. Representation decides whether the state can fit, communicate, and remain meaningful at low precision.

Within that stack, the matrix-aware renaissance is the central scientific development: Muon demonstrates that a cheap polar transform can be powerful at scale, while SOAP, SPlus, and related methods show the continuing value of variance and covariance structure. Memory-reduced, schedule-free, and quantized methods are not side stories; they determine which geometry is operationally possible.

There is therefore no single optimizer to replace AdamW. The more consequential transition is from choosing a brand name to designing and evaluating a stack. The winning optimizer will increasingly be the one that matches the parameter block, horizon, precision, and system budget of the run---and states those conditions clearly enough to reproduce.

\clearpage
\appendix
\section{Compact literature map}

\small
\begin{longtable}{P{0.16\textwidth}P{0.18\textwidth}P{0.36\textwidth}P{0.20\textwidth}}
\caption{Representative works used in this review. Evidence characterizes the cited claim, not the permanent status of the method.}
\label{tab:litmap}\\
\toprule
\textbf{Family} & \textbf{Work} & \textbf{Primary intervention} & \textbf{Evidence emphasis} \\
\midrule
\endfirsthead
\multicolumn{4}{l}{\footnotesize Table \thetable\ continued from previous page}\\
\toprule
\textbf{Family} & \textbf{Work} & \textbf{Primary intervention} & \textbf{Evidence emphasis} \\
\midrule
\endhead
\midrule
\multicolumn{4}{r}{\footnotesize Continued on next page}\\
\endfoot
\bottomrule
\endlastfoot

Temporal & AdEMAMix \citep{pagliardini2024ademamix} & Fast and slow gradient EMAs & Scaled, method-specific \\
Temporal & MARS \citep{yuan2024mars} & Recursive variance-reduced momentum & Controlled and scaled \\
Temporal & Cautious \citep{liang2024cautious} & Gradient--momentum agreement mask & Broad empirical \\
Temporal & SPAM \citep{huang2025spam} & Spike clipping and periodic state reset & Broad empirical \\
Temporal / low precision & Stable-SPAM \citep{huang2025stablespam} & Historical norm tracking, clipping, and resets for 4-bit training & Controlled and scaled \\
Temporal & ADOPT \citep{taniguchi2024adopt} & Decouples current gradient from current denominator & Theory plus empirical \\
Robustness / low precision & S-Adam \citep{xu2026singularity} & Randomized probes and instability-dependent damping & Theory plus emerging empirical \\
\addlinespace
Spectral & Muon \citep{jordan2024muon,liu2025muonscalable} & Approximate polar factor of momentum matrices & Scaled \\
Spectral & Scion \citep{pethick2025scion} & Norm-constrained linear minimization oracles & Theory plus empirical \\
Whitening & SOAP \citep{vyas2024soap} & Adam in evolving Shampoo eigenbases & Controlled and scaled \\
Whitening & SPlus \citep{frans2025splus} & Stable whitening, shape scaling, averaging & Controlled \\
Hybrid matrix & COSMOS \citep{liu2025cosmos} & SOAP leading subspace, Muon residual & Controlled \\
Hybrid matrix & AdaMuon \citep{si2025adamuon} & Second moments after orthogonalization & Emerging \\
Hybrid matrix & NorMuon \citep{li2025normuon} & Rowwise variance normalization after polar step & Scaled, method-specific \\
Hybrid matrix & Mousse \citep{sun2026mousse} & Kronecker whitening before polar step & Emerging \\
Hybrid matrix & MuonEq \citep{chang2026muoneq} & Row/column equilibration before polar step & Emerging \\
Hybrid matrix & Muown \citep{lion2026muown} & Explicit row-norm control & Emerging \\
Hybrid matrix & AngularMuown \citep{hubler2026angularmuown} & Explicit control of angular step size & Emerging \\
\addlinespace
Memory & Adam-mini \citep{zhang2024adammini} & Blockwise second moments & Scaled \\
Memory & SlimAdam \citep{kalra2025slimadam} & SNR-guided aggregation of second moments & Controlled \\
Memory & APOLLO \citep{zhu2024apollo} & Random low-rank auxiliary state & Scaled \\
Memory & LDAdam \citep{robert2024ldadam} & Projected moments with state transport & Controlled \\
Memory & GaLore2 \citep{su2025galore2} & Scalable low-rank gradient projection & Scaled \\
Memory & Alice/RACS \citep{gong2025structured} & Low-rank eigen-Adam / row-column-scaled SGD & Controlled and scaled \\
Memory & SWAN \citep{ma2024swan} & Stateless instantaneous normalization/whitening & Scaled \\
Memory & SCALE \citep{glentis2025scale} & Column-normalized SGD, minimal momentum & Scaled \\
\addlinespace
Horizon & Schedule-Free \citep{defazio2024schedulefree} & Fast/averaged iterates without known endpoint & Broad empirical \\
Horizon & WSD \citep{wen2024wsd} & Indefinite stable trunk plus decay branch & Controlled and scaled \\
Horizon & Power Scheduler \citep{shen2024power} & Token- and batch-aware learning-rate scaling law & Controlled and scaled \\
Horizon & WSM \citep{tian2025wsm} & Checkpoint merging as implicit decay & Controlled \\
Horizon & SF-NorMuon \citep{apte2026sfnormuon} & Schedule-free neuron-normalized spectral update & Emerging \\
\addlinespace
Low precision & COAT \citep{xi2024coat} & FP8 state and activations & Scaled \\
Low precision & SOLO \citep{xu2025solo} & 2--3 bit log-quantized states & Controlled and scaled \\
Low precision & 8-bit Muon \citep{gupta2025quantmuon} & Blockwise momentum quantization & Scaled, method-specific \\
Low precision & ECO \citep{nikdan2026eco} & No master weights; quantization error feedback & Emerging \\
Low precision & State staleness \citep{topollai2026statequant} & Analysis and resets for recursively quantized EMAs & Emerging \\
Low precision & STQuant \citep{liu2026stquant} & Layer- and time-dependent optimizer-state precision & Emerging \\
Quantization-aware pretraining & OSP \citep{park2025outliersafe} & Optimizer and parameterization choices that suppress activation outliers & Scaled, method-specific \\
\end{longtable}

\normalsize
\bibliographystyle{unsrtnat}
\bibliography{references}

\end{document}